\documentclass[oribibl]{llncs}

\usepackage{url}
\usepackage{epsfig}
\usepackage{program}
\usepackage{array}

\newcommand{\la}{\leftarrow}
\newcommand{\LA}{\Leftarrow}
\newcommand{\ra}{\rightarrow}

\newcommand{\ind}{\qquad\tab}

\renewcommand{\C}[1]{\mathcal{#1}}

\newcommand{\smodels}{{\sc smodels}}
\newcommand{\lparse}{{\sc lparse}}
\newcommand{\eclipse}{$ECL^iPS^e$ }

\begin{document}

\title{Logic Programming Approaches for Representing and Solving
Constraint
Satisfaction Problems: A Comparison}

\author{Nikolay Pelov \and Emmanuel De Mot \and Marc Denecker
\institute{Department of Computer Science, K.U.Leuven \\
Celestijnenlaan 200A, B-3001 Heverlee, Belgium \\
E-mail: \email{$\{$pelov,emmanuel,marcd$\}$@cs.kuleuven.ac.be} 
}}

\maketitle

\begin{abstract}
Many logic programming based approaches can be used to describe and
solve combinatorial search problems. On the one hand there is
constraint logic programming which computes a solution as an answer
substitution to a query containing the variables of the constraint
satisfaction problem. On the other hand there are systems based on
stable model semantics, abductive systems, and first order logic model
generators which compute solutions as models of some theory.  This
paper compares these different approaches from the point of view of
knowledge representation (how declarative are the programs) and from
the point of view of performance (how good are they at solving typical
problems).
\end{abstract}

\section{Introduction}

Consistency techniques are widely used for solving finite domain
constraint satisfaction problems (CSP) \cite{clp:Tsang}. These
techniques have been integrated in logic programming, resulting in
finite domain constraint logic programming (CLP) \cite{clp:cslp}.  In
this paradigm, a program typically creates a data structure holding
the variables of the CSP to be solved, sets up the constraints and
uses a labelling technique to assign values to the variables. The
constraint solver uses consistency techniques to prune the search.
This leads to a rather procedural programming style. Moreover, the
problem description is not very declarative because the mapping
between domain variables and their value has an indirect
representation in a term structure.

In this paper, we compare CLP and three computational paradigms
allowing problem solving based on more declarative representations.  A
common feature of these approaches is that the relation between the
CSP variables and their values is encoded as a predicate or function
relating identifiers of the CSP variables with their value. E.g. in
the graph coloring problem, the predicate relates node numbers with
colors.  This representation allows for a more natural declarative
representation of the problem.

One approach is specification in first order logic. As pointed out in
\cite{clp:Mackw92}, one can represent a CSP as a first order logic
theory such that (part of) its models correspond to the solutions of
the CSP.  Hence first order model generators such as SEM
\cite{mg:Zhang95} can be used to solve such problems.

The two other approaches use extensions of logic programming.
Recently, a logic programming paradigm based on stable model semantics
\cite{sem:SMS} has emerged. Niemel\"a \cite{mg:smodels} proposes it as
a constraint programming paradigm, Marek and Truszczy\'nski
\cite{sem:MT99} introduce Stable Logic Programming and Lifschitz
\cite{sem:Lif@ICLP99} proposes Answer Set Programming. As described in
\cite{sem:MT99}, the methodology of these approaches is to encode a
computational problem by a logic program such that its stable models
represent the solutions. A number of efficient systems for computing
stable models have been developed.
Of these, Niemel\"a's \smodels\ \cite{mg:NS96,mg:smodels} is considered one
of the most performant systems.

Abduction \cite{abd:ALP} uses a similar predicate representation for
the relation between the identifiers of CSP variables and their value.
This predicate is declared to be open or abducible.  Constraining this
relation to be a solution, an abductive system will return models of
the abducible which are solutions of the CSP. 

We use some typical CSP problems to compare the merits of the various
approaches. One experiment is in graph coloring. We have compared the
representation and the performance of CLP with the three other
approaches in a sequence of experiments where the size of the graph
increases and the number of colors remains constant. Another
experiment is the n-queens problem where both the domain size and the
number of constraints increases with increasing problem size. We also
report on experiments using CLP, stable logic programming and
abduction for solving a complex real world scheduling problem. For
each different system, we have tried to use any special features
provided by it.

In Section \ref{sec:kr-csp} we review in more detail the various
approaches and systems, focusing mainly on the knowledge
representation aspects. Section \ref{sec:exp} reports on the
experiments and we conclude in Section \ref{sec:concl}.

We are not aware of any previous work which compares this wide range
of logic based systems for their suitability in solving CSP problems.
Mackworth \cite{clp:Mackw92} explores the space of possible CSP
formalizations but assesses neither the quality from point of view of
knowledge representation nor the performance of actual systems. Also,
approaches based on stable model semantics and abduction are not
included in his work. This paper is an extension and revision of
\cite{abd:comp} which focuses more on the formal relations between the
declarative specifications of the problems on the different systems.

One more problem which uses aggregate functions is included in the
present paper. So is an additional experiment for finding all
solutions of the n-queens problem. Finally, some comments from the
authors of the different systems were taken into account.

\section{Formalisms and Systems} \label{sec:kr-csp}

A {\em constraint satisfaction problem} (CSP) is usually defined as a
finite set of {\em constraint variables} $\C{X} = \{X_1,\ldots,X_n\}$
(the variables of the CSP), a finite domain $D_i$ of possible values
for each variable $X_i$, and a finite set of {\em constraint
  relations} $\C{R}$ where each $r\in \C{R}$ is a constraint between a
subset of the set $\C{X}$ of variables.  A {\em solution} is an
instantiation of the variables of $\C{X}$ which satisfies all the
constraints in $\C{R}$.

\subsection{Constraint Logic Programming} \label{sec:clp}
Constraint logic programming (CLP) \cite{clp:survey} is an extension
of logic programming where some of the predicate and function symbols
have a fixed interpretation over some subdomain (e.g. finite trees or
real numbers). Special purpose constraint solvers are integrated with
a logic programming system for efficient reasoning on these
symbols. This results in a very expressive language which can
efficiently solve problems in many domains. 

Van Hentenryck \cite{clp:cslp} pioneered the work on finite domain
constraint logic programming, CLP(FD), by introducing domain
declarations for the logic variables and integrating consistency
techniques as part of the SLD proof procedure.
A CLP(FD) system supports standard arithmetic relations ($=,\neq,<$)
and functions ($+,-,*$) on the natural numbers.
A typical formulation of the n-queens problem is as follows:
\begin{program}
\ind queens(N,L)\la
\ind length(L,N),
     domain(L,1,N),
     constrain\_all(L),
     labeling(L). \untab
constrain\_all([]).
constrain\_all([X\origbar Xs]) \la
\ind constrain\_between(X,Xs,1)
     constrain\_all(Xs). \untab
constrain\_between(X,[],N).
constrain\_between(X,[Y\origbar Ys],N) \la
\ind safe(X,Y,N),
     N_1\; is\; N+1,
     constrain\_between(X,Ys,N_1).\untab
safe(X_1,X_2,D)\la
\ind  X_1\neq X_2, abs(X_1-X_2)\neq D. \untab
\end{program}

Executing the query $queens(n,L)$ first creates a list $L$ with $n$
variables where the $i^{\rm th}$ variable gives the column position of
the queen on row $i$. Then the constraints expressed with the $safe/3$
predicate are added by using two nested recursive predicates. Such
procedural code for setting up constraints and the encoding of the
solution in a large data structure results in a rather procedural
style which is typical for the CLP approach.

\subsection{First Order Logic: Model Generation} \label{sec:modgen}

The most elegant solution for the n-queens problem is using many
sorted first order logic and first order model generation. Systems
like FINDER and SEM \cite{mg:Zhang95} are examples.
One can introduce functions (with the sorts of their domain and range)
and predicates (with the sorts of their domains and the sort $bool$ as
range). In addition, functions can be restricted to be injective,
bijective,~\ldots This allows to express the n-queens problem very
concisely as:
\begin{program} 
\ind D=\{1..n\}
~
pos: D\ra D\quad (bijection)
~
abs(pos(X_1)-pos(X_2))\neq X_2-X_1 \la X_1<X_2.
\end{program}
The first line declares $D$ as a sort with interpretation consisting
of the set of integers $1$ to $n$. The following line introduces the
function $pos/1$ as a bijection from $D$ to $D$. Hence, the range of
the function is a permutation of its domain. This function represents the
column positions of the queens.
The only remaining constraint is that queens have to be on different
diagonals. This is expressed by the formula on the third line using
the predefined functions $abs/1$ and $-/2$. Due to symmetry, one need
only to verify the constraint for pairs of queens $X_1,X_2$ such that
$X_1<X_2$.

Solutions are given by the interpretation of the $pos/1$ function in the
models of this theory. In principle, this approach is applicable on
any CSP problem by representing the CSP variables by logical
constants. However, in most cases, CSP variables are just an
encoding of some attribute of a set of first order objects, such as
the position of a queen or the color of a node in a graph. In such
cases, there is no need to introduce the CSP variable. The attribute
can be represented directly as a function or predicate on these
objects (e.g. $pos$).

As the domains of all sorts are finite, SEM first computes the
grounding of the theory and then uses backtracking combined with
various inference and simplification rules to guide the search for
models \cite{mg:Zhang95}.

\subsection{Stable Logic Programming}

In \cite{mg:smodels}, Niemel\"a proposes logic programming with the
stable model semantics \cite{sem:SMS} as a constraint logic
programming paradigm. The underlying idea is to represent a problem as
a set of rules, each rule being the declarative expression of a piece
of knowledge about the problem domain and such that the stable models
of the whole program are constrained to be solutions of the problem.

The \smodels\ system \cite{mg:NS96} is an efficient implementation of
the stable model semantics. It works with propositional rules and a
special pre-processing program is used for grounding strongly range
restricted logic programs. The implementation combines bottom-up
inference with backtracking search and employs powerful pruning
methods. A recent extension of the system \cite{mg:sm2} introduces
choice rules:
\begin{program}
\ind l\; \{l_1, l_2, \ldots l_n\}\; u \la B.
\end{program}
where $l_1, l_2, \ldots l_n$ are literals. The semantics of such a
rule is that if the body $B$ is true then at least $l$ and at most $u$
literals among $l_i$ should be true in a stable model of the program.

Following \cite{mg:smodels} and \cite{mg:sm2}, the program for the
n-queens problems can be formulated as:
\begin{program}
\ind d(1..n).
~
1 \; \{pos(X,Y):d(Y) \} \; 1 \la d(X).
1 \; \{pos(X,Y):d(X) \} \; 1 \la d(Y).
~
\la d(X_1), d(Y_1), d(X_2), d(Y_2), pos(X_1,Y_1), pos(X_2,Y_2), 
\ind X_1 < X_2, X_2 - X_1 = abs(Y_1 - Y_2). \untab
\end{program}
Solutions are given by the $pos(i,j)$ atoms in the stable models of
the program. The first line defines that $d/1$ is a domain with
elements $1..n$ with $n$ the size of the board.  The first choice rule
is used to define the solution space of the problem by stating that
for each $X$ in the domain $d(X)$, there exists exactly one $Y$ such
that $pos(X,Y)$ is true. The colon notation denotes an expansion of
$pos(X,Y)$ for every value of $Y$.  Similarly, the second choice rule
expresses that there is exactly one queen on each column.  The last
rule defines the final constraint of the problem: no two queens on the
same diagonal. Again, the ``$<$'' constraints in these rules eliminate
instances which are redundant due to symmetry.  The main difference
with the first order logic specification is that the mapping between
queens and their position is now represented by a predicate. Declaring
that this predicate represents a bijective function is succinctly
expressed by the two choice rules.

\subsection{Abduction}

{\em Abductive logic programming} \cite{abd:ALP} extends the logic
programming paradigm with abductive reasoning. An abductive logic
program has three components: (1) a logic program $P$, (2) a set of
predicates $A$ called abducibles or open predicates, and (3) a set of
integrity constraints $I$. The abducibles are predicates not defined
in the program. The task of an abductive system is to find a set
$\Delta$ of ground abducible atoms such that the integrity constraints
are true in the logic program consisting of $P\cup\Delta$; formally:
$P\cup\Delta\models I$.

Kakas and Michael proposed an integration of CLP and an abductive
logic programming system \cite{abd:ACLP@ICLP95}.
Originally, it was defined only for definite
programs and integrity constraints and in \cite{abd:ACLP@JLP00} it was
extended to deal with negation as failure through abduction in a
similar way as in \cite{abd:nf}.
One restriction of ACLP is that integrity constraints need to be of
the form $\la a(\bar{X}),B$, where $a$ is an abducible. As we will
see, this forces sometimes to reformulate some constraints by an
additional recursion. Such restrictions are not present in SLDNFAC
\cite{abd:SLDNFA-CLP}, a more recent integration of an abductive
system with CLP that is based on the more general abductive procedure
SLDNFA \cite{abd:SLDNFA}.

The SLDNFAC system uses ID-Logic \cite{sem:ID-logic} as specification
language which is transformed into an abductive logic program by using
a Lloyd-Topor transformation. The specification of the n-queens
problem is:
\begin{program}
\ind d(1..n).
~
open\_function(pos(d,d)).
~
Y_1\neq Y_2 \wedge X_2-X_1\neq Y_2-Y_1\wedge X_2-X_1\neq Y_1-Y_2
\ind  \LA pos(X_1,Y_1) \wedge pos(X_2,Y_2) \wedge X_1<X_2.
\untab
\end{program}
The first line of the program defines $d/1$ as a domain predicate with
the integers $1..n$ as elements (defining rows and columns). The next
line states that the predicate $pos/2$ represents an open function in
the defined domain. It is used to represent the column position of a
queen in a row. Finally there is a constraint saying that two queens
can not be on the same column and diagonal. This representation is
almost identical to the FOL specification of section
\ref{sec:modgen}. The main difference is that the open function is
represented by a predicate.

As mentioned, ACLP does not allow function declarations. Consequently,
the fact that $pos$ predicate represents a function must be
expressed by explicit constraints. A standard way to axiomatize that
the abductive predicate $pos(X,Y)$ should be true for each $X$ in the
domain $d(X)$ is by using the following rule and integrity constraints:
\begin{program}
\ind has\_pos(X)\la d(Y),pos(X,Y).
\la d(X), not~has\_pos(X).
\end{program}
Unfortunately, the integrity constraint does not satisfy the ACLP's
restriction that at least one positive abductive atom should occur in it.
Hence, these axioms have to be reformulated using a recursive program
which generates a position for each queen. The specification for the
ACLP system is:
\begin{program}
\ind A = \{pos/2\}
~
     problem(N) \la nqueens(N,N).
     nqueens(0,N).
     nqueens(X,N)\la X > 0,~Y~in~1..N,~pos(X,Y),
     \ind X_{next}~is~X-1,~nqueens(X_{next},N).\untab
~
     attack(X_1,Y_1,X_2,Y_2) \la Y_1 = Y_2.
     attack(X_1,Y_1,X_2,Y_2) \la Y_1 + X_1  = Y_2 + X_2.
     attack(X_1,Y_1,X_2,Y_2) \la Y_1 - X_1  = Y_2 - X_2.
~
     \la pos(X_1,Y_1),~pos(X_2,Y_2),~X_1 < X_2,~attack(X_1,Y_1,X_2,Y_2).
\end{program}
The $n$-queens problem is solved by solutions of the abductive query $\la
problem(n)$.  The ACLP representation is in the middle of the
declarative FOL representation and the more procedural CLP
representation.

\section{Experiments} \label{sec:exp}

\subsection{The Systems}

The finite domain CLP package is the one provided with \eclipse version
4.2. 

Both abductive systems, ACLP \cite{abd:ACLP@JLP00} and SLDNFAC,
\cite{abd:SLDNFA-CLP} are meta interpreters written in Prolog, running
on \eclipse version 4.2 and making use of its finite domain library.
For all these systems, a search strategy which first selects variables
with the smallest domain which participate in the largest number of
constraints was used.

The model generator SEM version 1.7 is a fine tuned package written in
C. \smodels\ version 2.25, the system for computing stable models, is
implemented in C++ and the associated program used for grounding is
\lparse\ version 0.99.54. All experiments have been done on the
same hardware, namely Pentium II.

\subsection{Graph Coloring}

\begin{figure*}[htbp]
\begin{center}
  \psfig{file=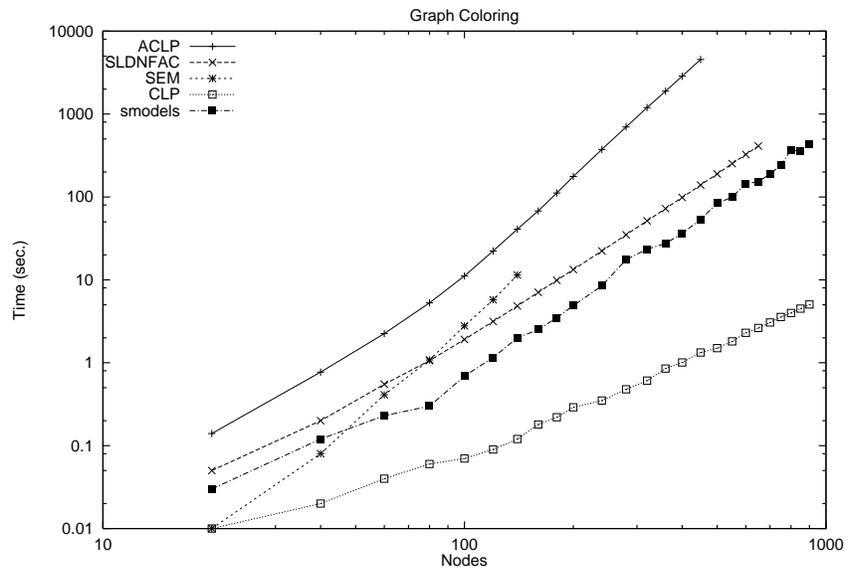,width=.9\textwidth}
\end{center}
\caption{Graph coloring} \label{fig:gc}
\end{figure*}

Our first experiment is done with 4-colorable graphs. We used a graph
generator\footnote{The graphs have been generated with the following
parameters: 0, 13, 6, n, 4, 0.2, 1, 0 where n is the number of
vertices. Graph-coloring problems generated with these parameters are
difficult.}  program which is available from address
\url{http://web.cs.ualberta.ca/~joe/Coloring/Generators/generate.html}.
We applied the systems in a sequence of experiments with graphs of
increasing size and constant number of colors. We have modified only
one parameter of the problem namely the number of vertices. Figure
\ref{fig:gc} gives the results of solving the problem with the
different systems.  Both axes are plotted in a logarithmic scale. On
the x-axis we have put the number of vertices.  Not surprisingly, CLP
is the fastest system. The times for \smodels\ is second best on this
problem. We assume it is in part because of the very concise
formulation. Using the so called technique of rules with exceptions
\cite{mg:smodels}, the two rules needed to describe the space of
candidate solutions also encode the constraint that the color is a
function of the vertex.  Hence there is only one other rule, namely
the constraint that two adjacent vertices must have a different
color.  The difference with CLP is almost two orders of magnitude for
the largest problems. The times reported for \smodels\ do not include
the time for grounding the problem, these times only consist of a
small part of the total time.  Grounding the problem for 650 nodes
takes only 10 seconds, whereas solving the problem takes over 100
seconds. SLDNFAC is slightly worse than \smodels. Although
meta-interpretation overhead tends to increase with problems size, the
difference with \smodels\ grows very slowly.  The model generator SEM
deteriorates much faster and runs out of memory for the larger
problems.  The fact that it grounds the whole theory is a likely
explanation. The difference with \smodels\ supports the claim that
\smodels\ has better techniques for grounding. ACLP performs
substantially worse than SLDNFAC and also deteriorates faster.  The
difference is likely due to the function-specification available in
SLDNFAC.
Contrary to ACLP, SLDNFAC exploits the knowledge that the abducible
encodes a function to reduce the number of explicitly stored integrity
constraints.

\subsection{N-Queens}

\begin{figure*}[htbp]
  \begin{center}
   \psfig{file=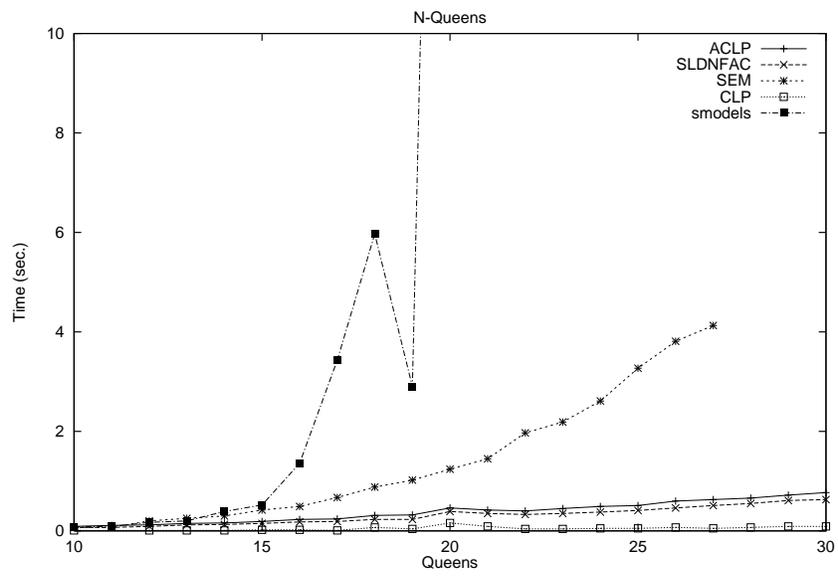,width=.9\textwidth}
  \end{center}
  \caption{N-queens: one solution} \label{fig:q}
\end{figure*}
\begin{figure*}[htbp]
  \begin{center}
   \psfig{file=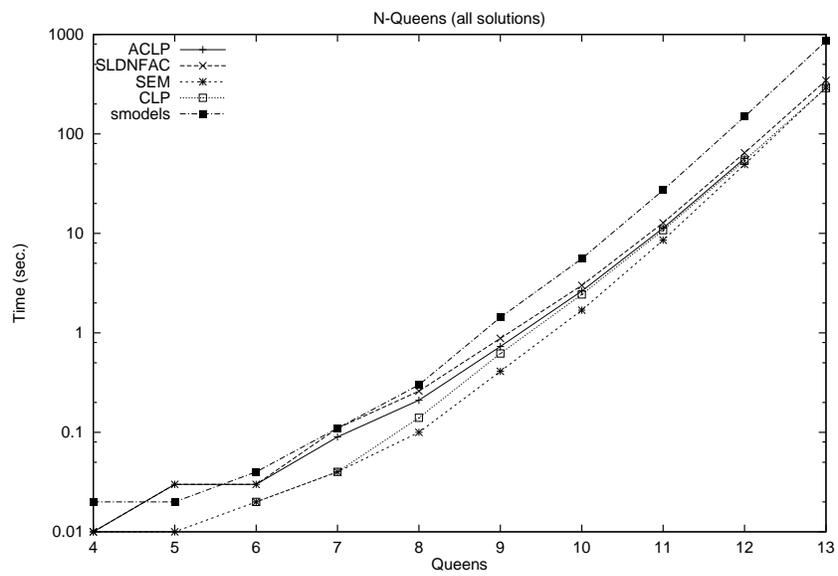,width=.9\textwidth}
  \end{center}
  \caption{N-queens: all solutions} \label{fig:qall} 
\end{figure*}

Figure \ref{fig:q} gives the running times for the different systems
for finding a first solution. Both axes are plotted on a linear scale.
The time consumed while grounding is again not included in the graph
(for 18 queens, half a second). Again, CLP gives the best results.
SLDNFAC is second best and, although meta-interpretation overhead
increases with problem size, deteriorates very slowly. ACLP is
third\footnote{The results with ACLP are substantially better than
  those in the previous paper \cite{abd:comp}. This is due to the
  removal of a redundant and time consuming complete consistency check
  after the processing of each new CLP constraint.}, with a small
difference, probably due to the lack of the function-specification
mentioned in the section above.  The next one is SEM. It runs out of
memory for large problems (it needs about 120MB for 27 queens). 
\smodels\ performs very poorly on this problem, in particular when
compared with its performance on the graph coloring problem. It is
well-known that to obtain good results for computing the first
solution for the n-queens problem, a good search heuristic is needed,
like the first fail principle used by the systems based on CLP. We
believe that the bad performance of \smodels\ is explained by the
absence of appropriate heuristics. This is confirmed by the much
better performance of the system in computing all solutions.

Figure \ref{fig:qall} gives the running times for finding all
solutions. The y-axis is plotted on a logarithmic scale.  The CLP,
ACLP and SLDNFAC systems are based on the same finite domain constraint
solver, so their convergence is not unexpected. Indeed, the abductive
system generates a constraint problem which is equivalent to the
problem generated by the CLP program and no backtracking occurs in the
abductive system. Hence, its overhead becomes ignorable. Also the SEM
system converges to the same performance as CLP (but runs out of
memory for big problems).
In this experiment, the \smodels\ system performs much better but is
still the slowest system. A likely reason for this is that the number
of propositional variables in the n-queens problem grows quadratically
with the problem size, in contrast with the graph coloring problem
where the number of variables grows only linearly (because of a
constant number of colors). Consequently, the grounding grows faster
for this problem. The CLP consistency techniques seem to be much less
sensitive to the domain size, and this carries over to the abductive
systems which reduce the problem to a CLP problem and then use the CLP
solver to search for the solution.

\subsection{A Real World Problem}

A Belgian electricity company has a number of power plants divided in
geographic areas. Each power plant has a number of power generating
units, each of which must receive a given number (usually 1 or 2) of
preventive maintenances with a fixed duration in the course of one
year. The computational problem is to schedule these maintenances
according to some constraints and optimality criteria.  Some of the
constraints are: some time slots are prohibited for maintenance for some
units; for each power plant, there is an upper limit on the total
number of units in maintenance per week for reasons of availability of
personnel; some of the maintenances are fixed in advance, \ldots The
objective of the problem is to find a schedule that maximizes the
minimal weekly reserve, which is the sum of the capacity of all units
not in maintenance minus the expected weekly peak load.

This is a rather difficult problem in several aspects. Firstly, the
specification uses aggregate expressions like cardinality and sum
(e.g. for each area, there is an upper limit to the total capacity for
units in maintenance per week). Only CLP, \smodels\ and SLDNFAC support
some form of aggregates and only these systems were used in our
experiment. Also, the search space is very large, as there are 56
maintenances to be scheduled in 52 weeks which makes about $56^{52}$
combinations\footnote{The maintenances with duration of more than one
  week cannot be scheduled in week 52, hence this number is only an
  upper approximation.}. The company provided a set of constraints for
which the optimal solution was known to have a minimal week reserve of
2100 (100\%). The three systems found correct schedules but none was
able to find this optimal solution.

This application was first considered in a context of a master's
thesis \cite{abd:tract-mt} and then reported in \cite{abd:Tractabel},
where a first attempt was done for integrating the SLDNFA proof
procedure with the CLP system ROPE \cite{clp:ROPE,clp:henk-phd}. This
early system needed 24 hours to reduce the problem to a constraint
store. Later on, in \cite{clp:henk-phd} several different direct
encodings in CLP of the problem were presented and compared. Recently,
\cite{abd:aggr} discussed an extension of the SLDNFAC system with
aggregate functions and this problem was used as a benchmark.

The first version of the \smodels\ system did not support aggregate
expressions. A more recent version of the system added a limited
support for rules with a body consisting of a single cardinality or
sum constraint \cite{mg:sm2} and allowed us to specify the problem.
However, these aggregate constraints cannot be used for computing the
sum or the cardinality of a set of atoms and we were not able to
express the optimization function. By setting increasing lower bounds
on the reserve capacity, branch and bound can be simulated manually.
It should be noted that, because of the very large size of the
problem, the specification of the problem in the \smodels\ system had
to be redesigned with special care in order to produce a ground
program not exceeding the limits of the system.

Table \ref{tab:tract} summarizes the results of executing the problem
with the different systems.  The first row ``Setup'' gives the time
used for pre-processing the problem specification. For the abductive
systems, this is the time for reducing the high-level specification to
a set of constraints.  For the \smodels\ system this is the time for
grounding the program. The rest of the rows give the times used by the
constraint solver to find a solution with the given quality.  The
results for CLP are taken from \cite{clp:henk-phd} for a standard
encoding of the problem\footnote{Without using global constraints,
like $cumulative$.} and the program was run under SICStus Prolog.

\begin{table}[htbp]
\begin{center}
\begin{tabular}{|l|r|>{\raggedleft}p{.55in}|r|r|}
\hline
Reserve & CLP    & SLDNFAC & \smodels \\
\hline
Setup &          &      45 & 36.4 \\  
\hline
1900  &          &    63.2 &  8.07 \\
2000  &     7.71 &    62.9 & $>$8h\\    
2010  &    25.85 &    63.8 & \\
2020  &    43.73 &    62.9 & \\ 
2030  &    57.28 &    63.0 & \\
2040  &    71.63 &   261.1 & \\
2050  & 26843.50 &   871.3 & \\
\hline
\end{tabular}
\caption{Power plant scheduling}
\label{tab:tract}
\end{center}
\end{table}

In the case of SLDNFAC, it can be seen in Table \ref{tab:tract} that
substantial progress was made. Rather than the 24h needed in the
earlier version \cite{abd:Tractabel}, the current SLDNFAC procedure
only needs 45 seconds for reducing the problem and about 15 minutes
for finding a solution of level 2050 (97.6\%). A solution with reserve
capacity of 2030 (96.5\%) was found in less than two minutes.  Note
that the timings for a solution with a reserve capacity of 1900 up to
2030 are similar. This is explained by the fact that in the five cases
the same solution with reserve capacity of 2030 was computed.  The
small differences in timings are due to noise in the measurements.
Strange enough, CLP deteriorates when it reaches a solution for a
reserve capacity of 2050 whereas the SLDNFAC solution does not. This
must be due to the fact that the constraint store built by the CLP
solution differs from the one built by the SLDNFAC solution. This is
accidental: in general, constraint stores constructed by a hand made
CLP program are more efficient than the ones computed by SLDNFAC. The
\smodels\ system needed 40 seconds for grounding and the best solution
we were able to find was 1900 (90.5\%) in 8 seconds. We did not find
better solutions in reasonable time.

\section{Conclusion} \label{sec:concl}

Finite domain CLP is widely accepted as an excellent tool for CSP
solving. However CLP programs have drawbacks from the point of view of
knowledge representation. As explained in Section \ref{sec:clp}, the
variables of the CSP have to be organized in a data structure and
``procedural'' code is required to create this data structure and to
set up the constraints. This level of indirection increases the
conceptual distance between the program and the problem and makes
programs less declarative.  Recently, several attempts have been made
to introduce formalisms allowing more declarative formalizations. They
are based on stable model semantics
\cite{sem:Lif@ICLP99,sem:MT99,mg:smodels} and on abduction
\cite{abd:ACLP@ICLP95,abd:ACLP@JLP00,abd:SLDNFA-CLP}. Although these
systems have an expressivity beyond what is needed to describe a CSP
(they address non-monotonic reasoning while CSP solving requires only
negation of primitive constraints), it is worthwhile to compare
these systems with CLP which is state of the art for CSP solving.
Because both stable models and abduction express solutions to problems
as models of their theory, we have also included first order model
generators in our study \cite{mg:Zhang95}. As argued in Section
\ref{sec:kr-csp}, these three approaches are better than CLP
from knowledge representation point of view, the formalizations are
more natural, more readable, conceptually closer to the problem, in
short they are more declarative than CLP programs.  Which one of the
three discussed mechanisms is the most declarative is likely a matter
of taste and familiarity.

Inevitably there is a price to be paid for these higher level
descriptions. None of the ``declarative'' systems experimented with
comes close to the performance level of CLP. This result holds
although the CLP system is not favored by the problem choice. Indeed,
in both graph coloring and n-queens problem, all constraints are
disequality constraints which are known to give little propagation.

Our experiments show that first order model generators do not scale
well and run out of memory for large problem instances even though the
size of the ground program is smaller compared to \smodels. We think
that this is not an inherent limitation of the approach but rather
that such systems were written with the goal of fast performance and
this is visible in our experiemnts. In contrast, \smodels\ runs in
linear space wrt the size of the grounding \cite{mg:NS96} and was able
to solve all problem sizes.
Of the two abductive systems, SLDNFAC supports a substantially richer
formalism and is performing slightly better than ACLP.  As the two
systems follow more or less the same strategy of top-down reduction of
integrity constraints and of forwarding the reduced ones to the CLP
solver and as both are implemented as a Prolog meta-interpreter, the
difference seems to be mainly due to the support of function
specifications.  The fact that the SLDNFAC meta-interpreter
outperforms SEM (a fine tuned C implementation) on both problems and
compares very well with the C++ implementation of \smodels\ (it is
much better on the n-queens problem while it reaches almost the same
performance on the graph coloring problem) suggest that its overall
strategy is the best one of the three systems for CSP solving. Also
the experiments with the large scheduling problem suggest this: the
setup time is acceptable and differences in search time seem to be due
to differences in the order of traversing the search space. While the
difference with CLP is substantial, a low level implementation or
compilation should be able to come close to the performance levels of
CLP, offering the best of both worlds: declarative problem
formulations and efficient execution. However, SLDNFA, the procedure
underlying SLDNFAC, is complex, hence building a direct implementation
is a hard task.
We believe the development of such a system is a worthwhile topic for
future research.

\section*{Acknowledgements}

Nikolay Pelov, Emmanuel De Mot and Marc Denecker are supported by the
GOA project LP+.
We want to thank Maurice Bruynooghe for his contribution on the topic
and anonymous referees for their useful comments.



\end{document}